\documentclass[journal]{IEEEtran}

\usepackage{algorithm}
\usepackage{algorithmicx}
\usepackage[noend]{algpseudocode} 

\usepackage{graphicx}
\usepackage{cite} 
\usepackage{times}
\usepackage{epsfig}
\usepackage{amsmath}
\usepackage{amssymb}
\usepackage{mwe}
\usepackage{acro}
\usepackage{amssymb}
\usepackage{xcolor,colortbl}
\usepackage{tabularx}
\usepackage{relsize}
\usepackage{pifont}
\usepackage{booktabs} 
\usepackage{multirow}
\usepackage{multicol}
\usepackage{adjustbox}
\usepackage{float}
\usepackage{url}
\usepackage{graphicx}
\usepackage[colorlinks,
            linkcolor=red,  
            anchorcolor=blue, 
            citecolor=green,   
            ]{hyperref}
\usepackage{amsmath,amsfonts,amssymb}
\usepackage{pifont}
\usepackage{graphicx}
\usepackage{setspace}
\usepackage{tocloft}
\usepackage{textcomp}
\usepackage{color,soul}
\usepackage{subcaption}
\usepackage{float}
\usepackage{multirow}
\usepackage{lineno}
\usepackage{makecell}
\usepackage{textcomp}
\usepackage{array}
\usepackage{listings}
\usepackage{mathptmx}
\usepackage{lineno}
\usepackage{epsfig}
\usepackage{times}
\usepackage{float}
\usepackage{rotating}
\usepackage{makeidx}
\usepackage{url}
\usepackage{multirow}
\usepackage{booktabs}
\usepackage{tabularx}
\usepackage{blindtext}

\DeclareAcronym{ROI}{
short=ROI,
long=region of interest,
}

\DeclareAcronym{IOU}{
short=IOU,
long=intersection over union,
}

\DeclareAcronym{cIOU}{
short=cIOU,
long=circle intersection over union,
}

\DeclareAcronym{DoF}{
short=DoF,
long=degrees of freedom,
}

\DeclareAcronym{CPL}{
short=CPL,
long=Center Point Localization,
}

\begin{document}
%
\title{Glo-In-One: Holistic Glomerular Detection, Segmentation, and Lesion Characterization with Large-scale Web Image Mining}
%
%
%

\author{Tianyuan~Yao,
Yuzhe~Lu,
Jun~Long,
Aadarsh~Jha,
Zheyu~Zhu,
Zuhayr~Asad,
Haichun~Yang,
Agnes~B.~Fogo,
Yuankai~Huo


\thanks{T. Yao, Y. Lu, Z. Zhu, A. Jha, Z. Asad, Y. Huo were with the Department of Computer Science, Vanderbilt University, Nashville, TN 37235 USA.}

\thanks{J. Long was with Big Data Institute, Central South University, Changsha, 410083, China.}

\thanks{A. B. Fogo and H. Yang were with the Department of Pathology, Vanderbilt University Medical Center, Nashville,
TN, 37215, USA.}

}


\maketitle

\begin{abstract}
The quantitative detection, segmentation, and characterization of glomeruli from high-resolution whole slide imaging (WSI) play essential roles in the computer-assisted diagnosis and scientific research in digital renal pathology. Historically, such comprehensive quantification requires extensive programming skills in order to be able to handle heterogeneous and customized computational tools. To bridge the gap of performing glomerular quantification for non-technical users, we develop the Glo-In-One toolkit to achieve holistic glomerular detection, segmentation, and characterization via a single line of command. Additionally, we release a large-scale collection of 30,000 unlabeled glomerular images to further facilitate the algorithmic development of self-supervised deep learning. The inputs of the Glo-In-One toolkit are WSIs, while the outputs are (1) WSI-level multi-class circle glomerular detection results (which can be directly manipulated with ImageScope), (2) glomerular image patches with segmentation masks, and (3) different lesion types. In the current version, the fine-grained global glomerulosclerosis (GGS) characterization is provided, including assessed-solidified (S-GGS, associated with hypertension-related injury), disappearing (D-GGS, a further end result of the SGGS becoming contiguous with fibrotic interstitium), and obsolescent (O-GGS, nonspecific GGS increasing with aging) glomeruli. To leverage the performance of the Glo-In-One toolkit, we introduce self-supervised deep learning to glomerular quantification via large-scale web image mining. The GGS fine-grained classification model achieved a decent performance compared with baseline supervised methods while only using 10\% of the annotated data. The glomerular detection achieved an average precision of 0.627 with circle representations, while the glomerular segmentation achieved a 0.955 patch-wise Dice Similarity Coefficient (DSC). In this paper, we develop and release an open-source Glo-In-One toolkit, a software with holistic glomerular detection, segmentation, and lesion characterization. This toolkit is user-friendly to non-technical users via a single line of command. 

The toolbox and the 30,000 web mined glomerular images have been made publicly available at  \url{https://github.com/hrlblab/Glo-In-One}.

\end{abstract}


{\noindent \footnotesize\textbf{*Corresponding Author:} Yuankai Huo,  \url{yuankai.huo@vanderbilt.edu} }

%
\IEEEpeerreviewmaketitle

\section{Introduction}

\begin{figure*}[t]
\begin{center}
 \includegraphics[width=0.94\linewidth]{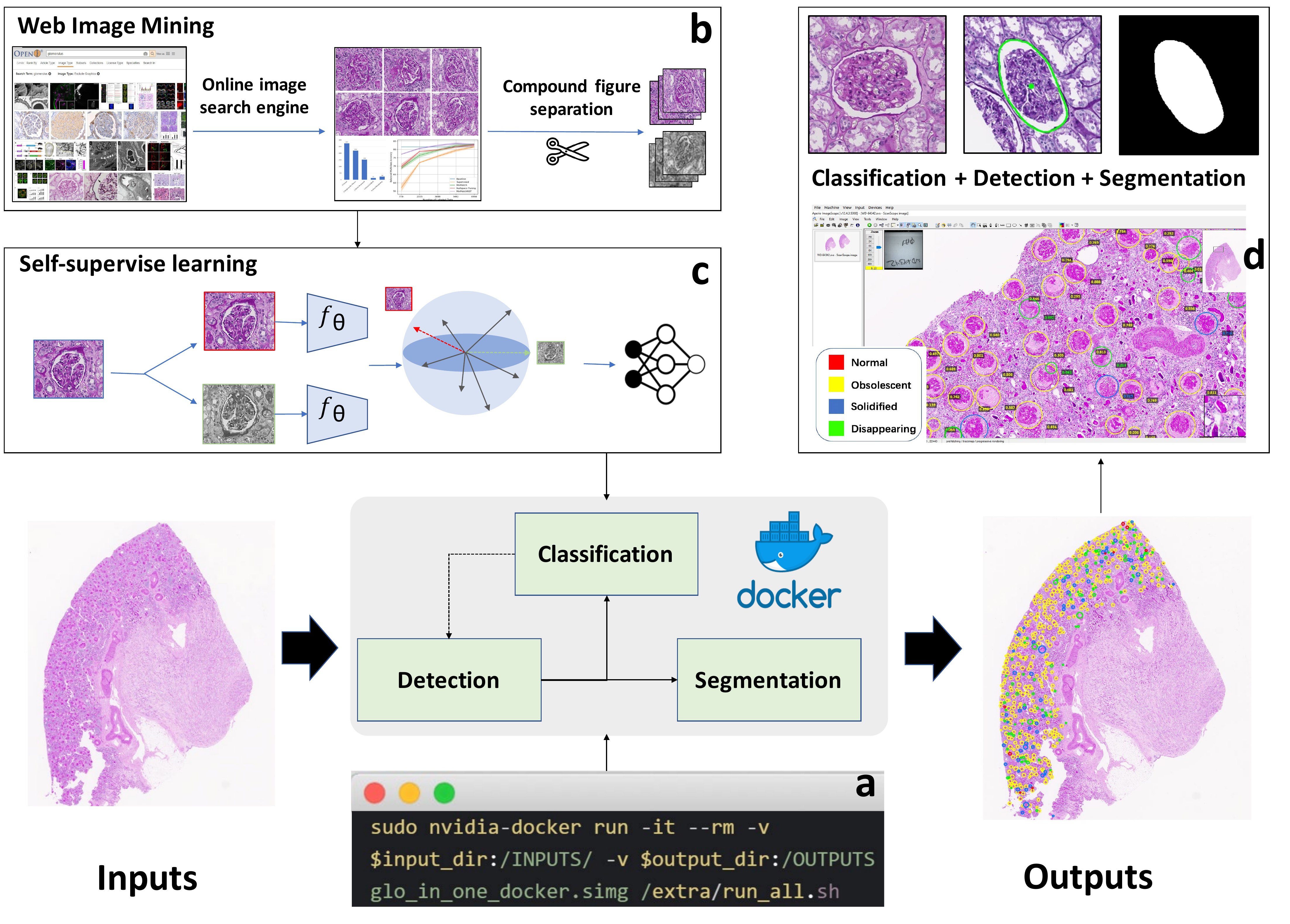}
\end{center}
   \caption{This figure presents the overview of the Glo-In-One toolkit. The inputs are raw WSIs, while the outputs are quantitative detection, classification, and segmentation results. Moreover, the results can be manipulated using the prevalent Aperio ImageScope software. The domain impact is to enable comprehensive glomerular quantification via one single command for non-technical users. The technical innovation is to introduce large-scale self-supervised deep learning via web image mining.}
  \label{fig:overview}
\end{figure*}  

\IEEEPARstart{G}{lomeruli} are central tufts of looped capillaries located in the center of the renal corpuscle~\cite{hoy2003stereological,kimmelstiel1936intercapillary}, which delivers blood and creates a large surface area optimized for renal filtration~\cite{murray2020histology,scott2015cell}. Identiﬁcation and characterization of glomeruli are essential for computer-assisted pathological diagnosis in digital renal pathology~\cite{santo2020artificial,rangan2007quantification}. However, such procedures are typically labor and resource intensive. The recent advances in whole slide imaging (WSI) and deep learning have led to an unprecedented opportunity to characterize glomerular lesions in a large-scale and fully automatic manner~\cite{barisoni2013digital,zeng2020identification,gallego2018glomerulus,nast2015morphology}. Typical high-resolution WSI images are in the order of gigapixels and are usually stored in multi-resolution pyramidal format~\cite{khened2021generalized}.

Historically, comprehensive computer-assisted glomerular quantification requires extensive programming skills to handle the heterogeneous and custom computational tools~\cite{kather2019deep, moen2019deep, mobadersany2018predicting, beck2011systematic, huo2021ai} . To bridge the gap of performing glomerular quantification for non-technical users, we develop the Glo-In-One toolkit to achieve holistic glomerular detection, segmentation, and characterization via a single line of command (Fig.~\ref{fig:overview}). 

For the glomerular detection~\cite{kawazoe2018faster,temerinac2017detection,bukowy2018region,maree2016approach}, we apply an anchor-free detection method with a circle representation~\cite{yang2020circlenet}. The simple circle representation is optimized for ball-shaped biomedical object detection with smaller degrees of freedom (DoF) in fitting and superior detection performance which is optimized for glomerular detection in renal pathology. Then, the Glo-In-One toolkit further segments and classifies each detected glomerulus with a pixel-level mask and fine-grained Global glomerulosclerosis (GGS) classification. GGS is a prevalent pathological phenotype for patients with both normal aging and kidney disease~\cite{abdelhafiz2010chronic,greenberg1997focal,lorbach2020clinicopathologic}. Fine-grained characterization of glomerular lesions~\cite{tan2020global}, is essential in nephropathology in order to support both scientific research and clinical decision making\cite{benchimol2003focal}. 
 
Lesion characterization involves the development of quantitative machine learning approaches (e.g., self-supervised deep learning) that characterize glomerular lesions typically require large-scale heterogeneous images~\cite{huo2021ai,ginley2017unsupervised,david2019applications}, which is resource extensive for individual labs. We first assess the feasibility of leveraging fine-grained GGS characterization via large-scale web image mining (e.g., from journals, search engines, websites) and self-supervised deep learning. Three types of GGS are assessed-solidified (S-GGS, associated with hypertension-related injury), disappearing (D-GGS, a further end result of the SGGS becoming contiguous with fibrotic interstitium), and obsolescent (O-GGS, nonspecific GGS increasing with aging).  We employ a SimSiam~\cite{chen2020exploring} network as the baseline method of self-supervised contrastive learning. By deploying our previously developed compound figure separation approach~\cite{yao2021compound}, we provide 30,000 unannotated glomerular images via web image mining to train the SimSiam network. From the results, the GGS fine-grained classification model achieved superior performance as compared with the baseline methods. We release such large-scale collection of unlabeled glomerular images to further facilitate the methodological development of self-supervised deep learning for the engineering community.

Our contribution is four fold:

\begin{itemize}
    \item We develop and release an open-source Glo-In-One toolkit: a containerized software with holistic glomerular detection, segmentation, and lesion characterization. The toolkit is user friendly to non-technical users via a single line of command, with standard WSI files as inputs.
    \item We assess the feasibility of mining large-scale web images for fine-grained GGS classification via contrastive learning. 
    \item We release a large-scale web-mined glomerular dataset with 30,000 unannotated glomeruli to facilitate the methodological development of self-supervised deep learning.
    \item The toolbox and the 30,000 web mined glomerular images have been made publicly available at  \url{https://github.com/hrlblab/Glo-In-One}
\end{itemize}

\section{Related Works}
The success of deep convolutional neural networks (CNNs) in recent years has spurred extensive research on their application in renal pathology~\cite{kolachalama2018association,hermsen2019deep,lu2017deep}. The innovations can be summarized as glomerular detection, segmentation, and characterization, which have further advanced digital diagnostic imaging, especially in WSIs technology~\cite{sarder2016automated,BUENO2020105273,kumar2020whole}.

\subsection{Glomerulosclerosis Classification}
Many studies have been conducted to classify different glomerular lesions using computer-aided approaches~\cite{marsh2018deep,zeng2020identification,ginley2020fully,chagas2020classification}. Specifically, Uchino et al.~\cite{uchino2020classification} trained deep learning algorithms to classify several renal pathological findings and achieved an AUC score of 0.983 for global sclerosis characterization. However, most algorithms are tested using in-house data and thus, their performances can be highly volatile in the face of a distribution shift~\cite{fernandez2011addressing}. Moreover, there are few, if any, studies that have developed deep learning approaches to classify globally sclerotic glomeruli into the following three fine-grained categories: obsolescent, solidified, and disappearing glomerulosclerosis. Such fine-grained characterization is challenging as the available data is rare and their distribution is highly imbalanced. For instance, the presence of obsolescent glomerulosclerosis is naturally much higher than solidified or disappearing glomerulosclerosis~\cite{cascarano2019innovative,pierpont2000unbalanced}, leading to the technical difficulty that is well known as the imbalanced classification problem~\cite{huang2016learning,zou2016finding,lopez2012analysis}.

\subsubsection{Self-supervised learning method}
Self-supervised learning represents a family of learning algorithms that could learn the hidden regularities from using data without manual annotations (e.g., unannotated renal pathological images)~\cite{hastie2009overview}. Such technology might alleviate the bottleneck of the scale of human annotation in healthcare as large-scale unannotated data could be available from clinical database. Moreover, the generalizability of the AI models on unseen testing data would be further improved by learning from a larger scale of training data, even without human annotations~\cite{celebi2016unsupervised}.

Recently, a new family of self-supervised representation learning, known as contrastive learning, has shown its superior performance in various computer vision tasks ~\cite{wu2018unsupervised,noroozi2016unsupervised,zhuang2019local,hjelm2018learning}. Using an approach where learning is achieved from large-scale unlabeled data, contrastive learning can learn discriminative features for downstream tasks. SimCLR~\cite{chen2020simple} maximized the similarity between images in the same category and repelled the representations of different category images. The ‘similarity’ is typically defined as the  cosine similarity between features~\cite{nguyen2010cosine} in the previous prior arts. Wu et al.~\cite{wu2018unsupervised} used an offline dictionary to store all data representation and randomly selected training data to maximize negative pairs. MoCo~\cite{he2020momentum} introduced a momentum design to maintain a negative sample pool instead of an offline dictionary. Such works demanded a large batch size in order to include sufficient negative samples. To eliminate the needs of negative samples, BYOL~\cite{grill2020bootstrap} was proposed to train a model with an asynchronous momentum encoder. Recently, SimSiam~\cite{chen2020exploring} was proposed to further eliminate the momentum encoder in BYOL, allowing  for less GPU memory consumption.

\subsection{Glomerular Detection}
The introduction of WSIs demonstrates a paradigm shift of computer-aided diagnosis (CAD) from visual inspection to a more accurate quantitative assessment. CAD, especially with recent advances in deep learning, offers an unparalleled ability to efficiently manage patients, accelerate diagnosis, and guide treatments in general health care. In renal pathology, the promising results of integrating CAD in chronic kidney diseases (CKD) and kidney transplantation have been shown from prior arts~\cite{becker2020artificial,huo2021ai}, especially with automatic glomerular characterization~\cite{marsh2018deep,barros2017pathospotter}. In renal pathology, properly distinguishing and characterizing different glomeruli from renal tissues is a key task in both corresponding scientific research and clinical practice. In recent years, such historically labor-intensive procedures have been advanced by modern deep learning techniques, via classification, detection, and segmentation approaches~\cite{lu2021data}. Several studies have shown the great accuracy by which CNNs are able to properly detect and localize glomeruli within WSIs. Similarly, CNNs have also been able to accurately segment glomeruli. Beyond identifying the regional and pixel location of glomeruli, deep learning algorithms have also been applied in CAD (e.g., pathological glomerular characterization), such as quantifying and characterizing lesion sub-types of glomeruli. For example, Gallego et al.~\cite{gallego2018glomerulus} used CNN-based classification, while Gloria et al.~\cite{bueno2020glomerulosclerosis} utilized semantic segmentation to achieve glomeruli detection in WSIs. More recently, Yang et al. proposed an anchor-free detection strategy using circle representation~\cite{yang2020circlenet} that is optimized for round glomeruli and demonstrates superior performance.

\subsection{Glomerular Segmentation}

An initial step for renal tissue assessment is the differentiation and segmentation of relevant tissue structures in kidney specimens as it is performing pixel-level quantification on the image~\cite{de2018automatic,hermsen2019deep,altini2020deep}. However, Segmentation on WSIs is challenging since the pixel-level annotation is computational extensive on the gigapixel high-resolution images~\cite{qaiser2019fast,gadermayr1708cnn}. Bouteldja et al investigated the concept of active learning for accurate segmentation accuracy~\cite{bouteldja2021deep} by performing a large number of 72,722 expert-based annotations, while Gadermayr et al has proposed a weakly supervised pipeline for segmenting renal glomeruli~\cite{gadermayr2017segmenting}. Other method, to achieve instance object level segmentation, has integrated fully convolutional networks with detection methods on WSIs(e.g., Mask-RCNN~\cite{he2017mask}), but there is still room to improve~\cite{cao2019gastric, lv2019nuclei}.Recently, Aadarsh et al.~\cite{jha2021instance} shows that the "detect-then-segment" two-stage segmentation approach yields more accurate results than the conventional single-stage instance segmentation tactics. Following this study, we propose a detect-classify-segment pipeline to achieve even more accurate glomerular segmentation results.

\section{Method}

\begin{figure*}
\begin{center}
\includegraphics[width=1\linewidth]{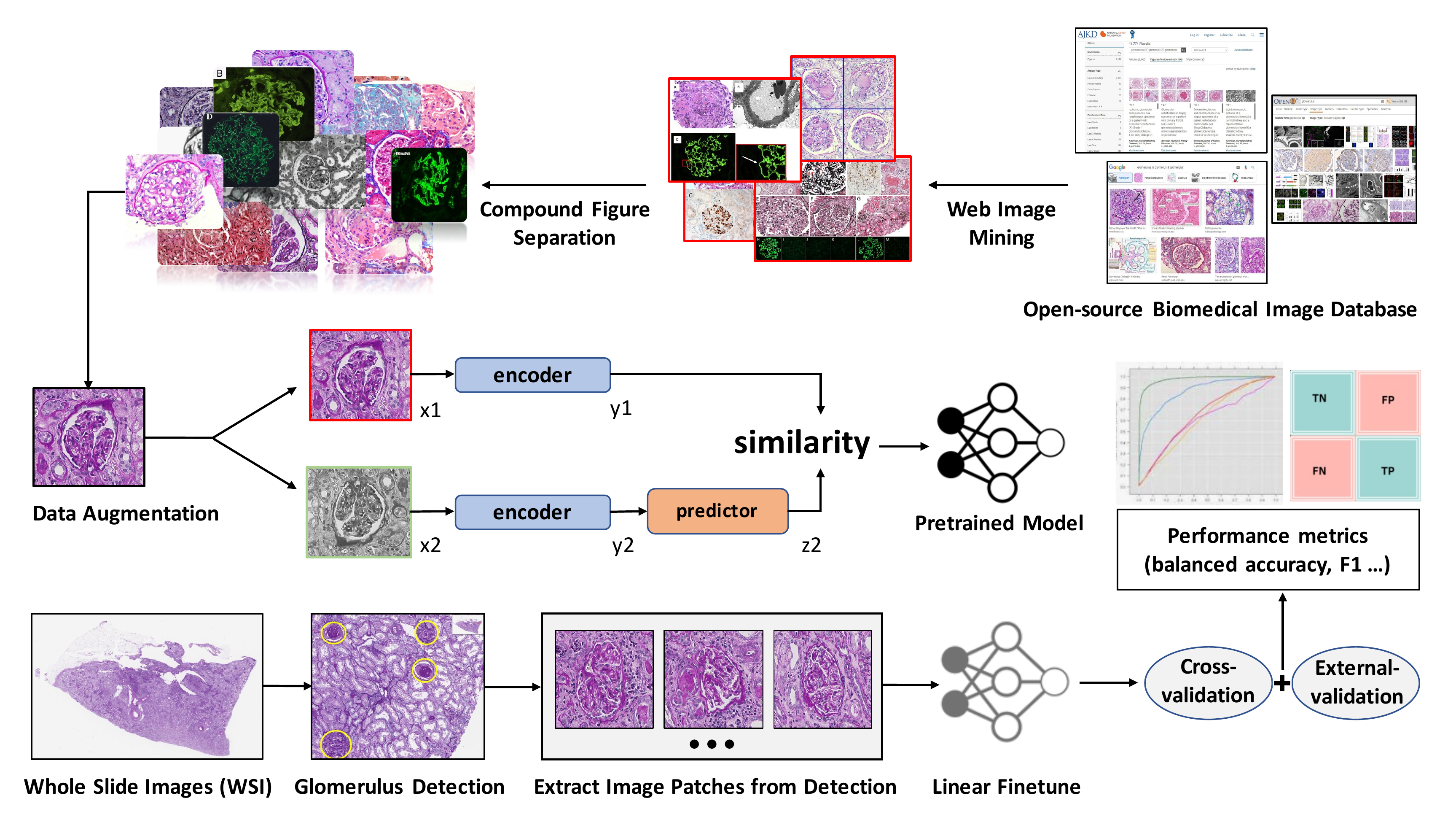}
\end{center}
   \caption{The self-supervised glomerular characterization pipeline is presented. Our approach utilizes the large-scale unannotated web-mined glomerular images via compound image separation to pretrain a contrastive learning model. Then, a smaller amount of labeled data, especially for rare GGS samples, are employed to finetune the pretrained model. The goal is to improve the glomerular classification via large-scale unannotated web-mined images and small-scale annotated in-house data.}
\label{fig:classifcation}
\end{figure*}

\begin{figure*}

\begin{center}
\includegraphics[width=1\linewidth]{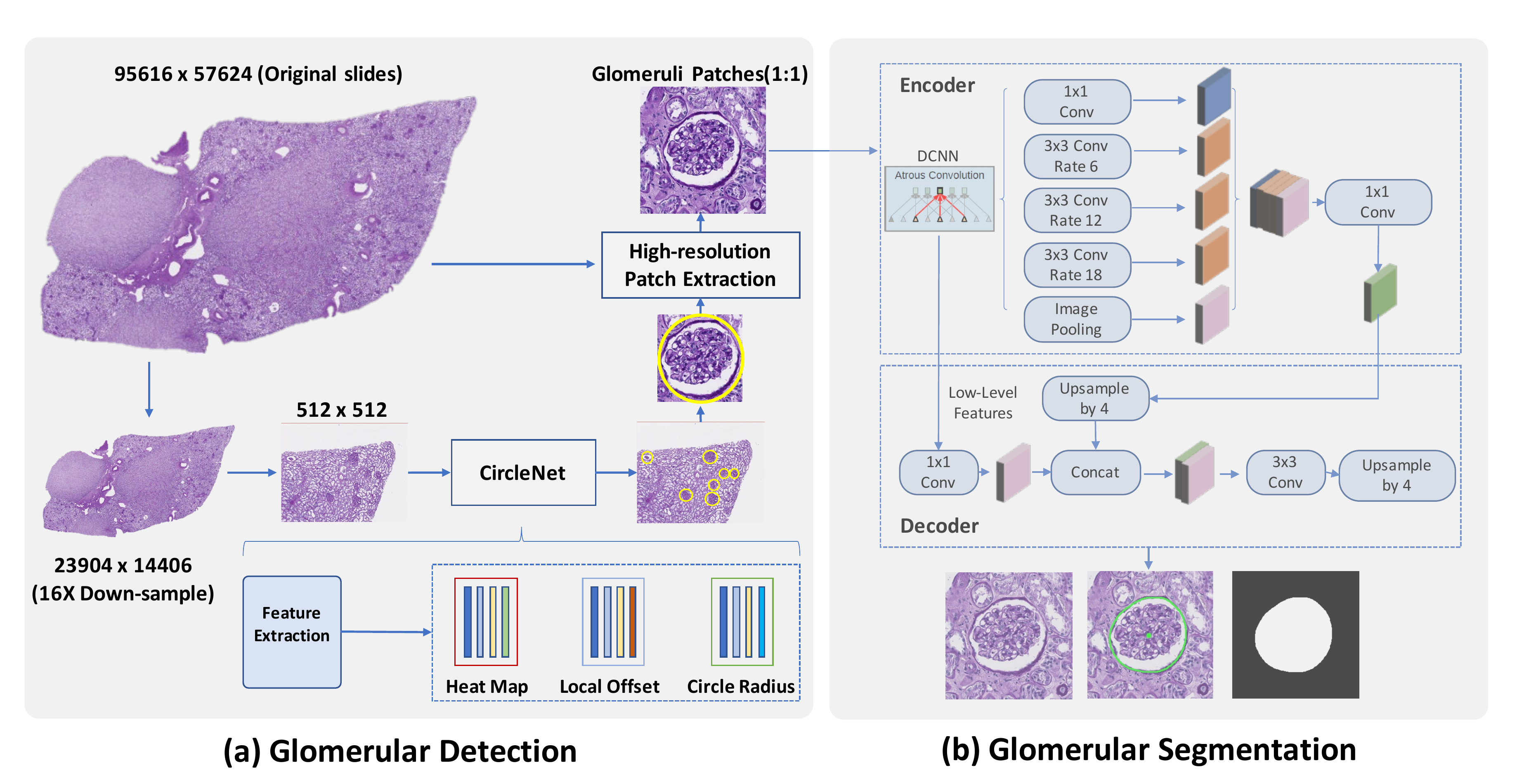}
\end{center}

\caption{(a) The detection framework with circle representation~\cite{yang2020circlenet} is presented, which is an optimized object detection method for glomeruli. The heatmap and local offset head determines the center point of the circle while the circle radius head determines the radius of the circle. (b) The DeepLab\_v3 backbone is employed in the "detect-then-segment" pipeline.}
\label{fig:detection}
\end{figure*}

Generally, the glomerular quantification in this study can be summarized into three major steps: (1) detection, (2) characterization, and (3) segmentation (Fig.~\ref{fig:classifcation} and~\ref{fig:detection}). First, CircleNet~\cite{yang2020circlenet,ethan2021circle} is employed to perform automatic glomerular detection. After the detection, we extract image patches that contain the detected glomeruli and add padding (50 pixels) around the patches. Then, all the patches are cropped and resized to 256$\times$256-pixel for downstream tasks using bilinear interpolation, while the information of the bounding box and actual size (0.25 $\mu$m per pixel) is saved in an XML file. For the task of classification, our self-supervised contrastive learning approach uses minimal human efforts (e.g., self-supervised learning) to achieve fine-grained GGS classification via large-scale web image mining. The procedure consists of two parts: (1) web data mining from biomedical databases via compound figure separation, and (2) self-supervised learning with downstream classification tasks. For the step of segmentation, we apply the DeepLab\_v3 model to perform the segmentation and generate binary masks.

\begin{figure*}
\begin{center}
\includegraphics[width=1\linewidth]{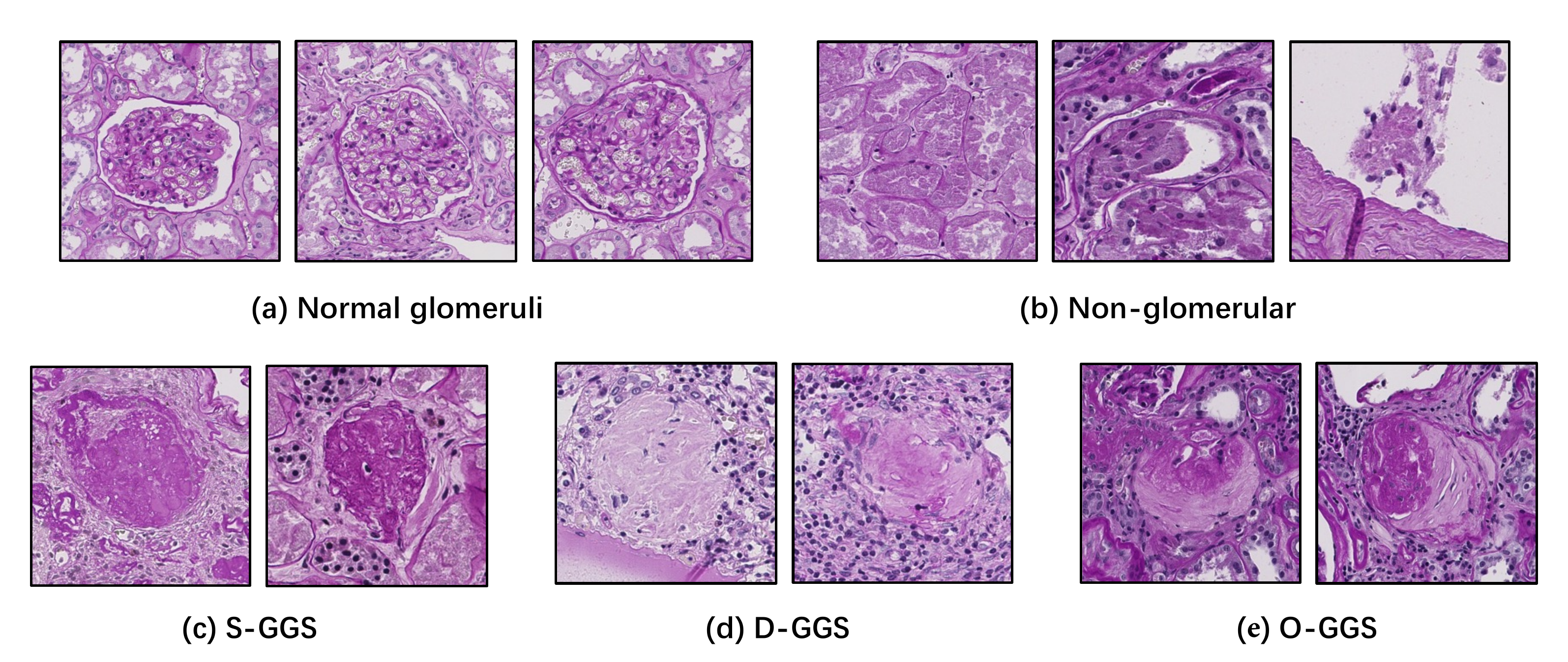}
\end{center}
   \caption{This figure shows fine-grained classes of glomerular images that are stained with PAS. (a) Normal glomeruli, (b) non-glomerular and (c)-(e) sclerosed glomeruli images. Three types of sclerosed glomeruli are (c) assessed-solidified (S-GGS, associated with hypertension-related injury), (d) disappearing (D-GGS, a further end result of the SGGS becoming contiguous with fibrotic interstitium), and (e) obsolescent (O-GGS, nonspecific GGS increasing with aging) glomeruli.}
   \label{fig:types}
\end{figure*}



\subsection{Glomerular Classification}
In the Glo-In-One toolkit, we performed a fine-grained GGS classification as the application (Fig.~\ref{fig:classifcation} and~\ref{fig:types}). First, all glomeruli images that were detected by Glo-In-One (details are provided the in the next section) were further refined as glomerular vs. non-glomerular images. Second, all confirmed glomeruli images were further classified into GGS vs. non-GGS images. Third, GGS images were characterized into fine-grained categories: (1) solidified (S-GGS, associated with hypertension-related injury), (2) disappearing (D-GGS, an end result of the S-GGS becoming contiguous with fibrotic interstitium), and (3) obsolescent (O-GGS, nonspecific GGS increasing with aging) glomeruli. The technical innovation of our method was introducing the contrastive learning as a self-supervised pretraining stage to enhance the glomerular classification performance. 

\subsubsection{Web Image Mining}
We propose to achieve large-scale unannotated glomerular images from online resources (e.g., open-access journals, NIH Open-i$^\circledR$~\cite{demner2012design} database, and search engines) in order to facilitate self-supervised contrastive learning (Fig.~\ref{fig:classifcation}). Briefly, we collect 10,000 compound figures with the keywords ``glomerular OR glomeruli OR glomerulus” through the NIH Open-i$^\circledR$ search engine. The details of web image mining are provided in~\cite{yao2021compound}.



However, the images from online resources are typically in compound figures (with multiple subplots), which cannot be directly used for self-supervised learning. Thus, we employ our previously developed compound image separation approach~\cite{yao2021compound} to detect, separate, and curate subplots to individual images for downstream learning tasks. Using the compound figure separation approach, we acquired over 30,000 unannotated glomerular images via large web image mining.

\subsubsection{Self-supervised Learning}
As opposed to traditional supervised learning methods, self-supervised learning~\cite{celebi2016unsupervised} refers to inferring underlying patterns from an unlabeled dataset without any reference to labeled outcomes or predictions (Fig.~\ref{fig:classifcation}). We employ the SimSiam network~\cite{chen2020simple} according to our previous pathological image analytical study using multiple contrastive learning methods~\cite{liu2021simtriplet}. The SimSiam network can be interpreted as an iterative process of two steps: (1) unsupervised clustering and (2) feature updates based on clustering (similar to K-means or EM algorithms)~\cite{chen2020exploring}. After training the SimSiam with large-scale mined images, a pretrained ResNet backbone is available as a so called feature encoder. Then, the output of the encoder is passed through a predictor which is again a shallow fully-connected network for the final classification results. 

\subsection{Glomerular Detection}
The glomerular detection function is implemented based on our previously proposed CircleNet method~\cite{yang2020circlenet,ethan2021circle}, which introduces a new bounding circle representations for glomerular detection (Fig.~\ref{fig:detection}). Since CircleNet achieved superior performance for glomerular detection compared with current benchmarks~\cite{yang2020circlenet,ethan2021circle}, we directly migrate CircleNet as the detection method in this study. The detection outcomes are saved in one XML file that contains the circle location, type, and the detection score for each detected object. The detection score is a score within 0 to 1, where a larger score indicates the stronger confidence to believe the detected object is a glomerulus. The XML format file can be loaded with pathology slide viewing software such as Aperio ImageScope.

\subsection{Glomerular Segmentation}
The glomerular segmentation function is implemented based on our previously proposed ``Detect-then-Segment" method~\cite{jha2021instance} (Fig.~\ref{fig:detection}). In segmentation, instead of directly apply segmentation approaches on lower resolution image patches, our ``Detect-then-Segment" two-stage design~\cite{jha2021instance} applies DeepLab\_v3~\cite{chen2017rethinking} on the detected high-resolution glomerular patches with minized information loss. Such design has been demonstrated to outperform the prevelent single-stage approaches~\cite{jha2021instance}. The DeepLab\_v3~\cite{chen2017rethinking} is employed as the segmentation backbone. DeepLab\_v3 utilizes atrous convolution which allows it to enlarge the field of view of filters to incorporate larger context. It thus offers an efficient mechanism to balance accurate localization (small field-of-view) and context assimilation (large field-of-view). 


\section{Experiments and Results}
\subsection{Glomerular Classification}

\subsubsection{Data}
\textbf{Model Pretrain.}In this study, we collected over 10,000 compound figures (each figure might contain multiple subplots) through the NIH Open-I$^\circledR$ search engine with the keywords “glomerular OR glomeruli OR glomerulus”. Then, our compound figure separation method~\cite{yao2021compound} was employed to separate compound images into individual images, which were further categorized to different modalities (e.g., light microscopy, florescent microscopy, and electron microscopy) as well as different stain types within the light microscopy. To curate all images, an automatic deep learning-based curator (detector) was trained using only a smaller scale annotated images dataset~\cite{bochkovskiy2020yolov4}. Then, the curator was run on all web mined images in order to drop the predicted glomerular figures under a detection score of 0.7. The 0.7 threshold was determined empirically by visual inspection. Eventually, more than 30,000 glomerular images (subplots) passed the curation. The number and size of the image patches used for different learning strategies are described in Experimental Design.

\begin{table}[h]
\centering{}
\caption{Distribution of subclasses in glomerular classifier}
\begin{tabular}{c|cc|c}
\toprule
Type  & Quantity & Percentage & Total \\
\midrule
Normal                 & 3,617     & 16\%    &        \\
obsolescent glomeruli  &6,647      & 30\%    &        \\
solidified glomerul    & 735       & 3\%     &        \\
disappearing glomerul  &459        & 2\%     &        \\
Non-glomerular         &10,619     & 49\%    & 22,077  \\
\bottomrule
\end{tabular}
\text{*As mentioned, the collected dataset is naturally unbalanced. }
\label{table:data}
\end{table}

\textbf{Cross Validation.} The classification models were trained on 22,077 images extracted from WSIs using the EasierPath semi-manual annotation software~\cite{zhu2020easierpath}. The WSIs were acquired from 157 patients, whose tissues were routinely processed and paraffin-embedded, with 3 $\mu$m thickness sections cut and stained with Periodic acid–Schiff (PAS). The WSIs were acquired at 40× with 0.25 $\mu$m per pixel.

All glomeruli were manually annotated with 5 classes, including 3,617 normal glomeruli, 6,647 globally obsolescent glomeruli, 735 global solidified glomeruli, and 459 global disappearing glomeruli, and 10,619 non-glomerular images (Table \ref{table:data}).  The images were resized to 224$\times$224-pixel for training, validation, and testing. The data were de-identified, and studies were approved by the Institutional Review Board (IRB) at VUMC. 

Additionally, 120 glomerular patches were derived from five WSIs of the kidney tissue which respectively from different patients and were utilized as detection test set in our automatic detection experimentation. The details of the data selection, acquisition and clinical background were provided in~\cite{ethan2021circle}.

For the segmentation experiment, we formed a cohort with 704 training, 98 validation and 147 internal testing images images which respectively extracted from 42, 7 and 7 WSIs using our detection method. Meanwhile, a group of 385 glomerulus pathologies from 5 external biopsy samples was used as external testing data. Both GGS and none-GGS glomerulus patches were manually annotated and the segmentation masks were traced by a experienced pathologist.

\textbf{External Validation.} In addition to our in-house datasets, a publicly available glomeruli dataset \cite{bueno2020data} was also employed as an external validation cohort. The dataset consisted of 2,340 images containing a single glomerulus. Among them, 1,170 glomeruli were normal, while 1,170 glomeruli were sclerosed. 

To further evaluate our method, the model was trained on both 5-classes and 2-classes (binary) designs. In the binary setting, (1) three types of sclerosed glomeruli (S-GGS, D-GGS, O-GGS) were merged as a single positive class; (2) the normal type was kept as the negative class; (3) 10,619 non-glomerular images were discarded to ensure consistency with the definition in \cite{bueno2020data}.

\subsubsection{Experimental Design}

\begin{figure*}
\begin{center}
\includegraphics[width=1\linewidth]{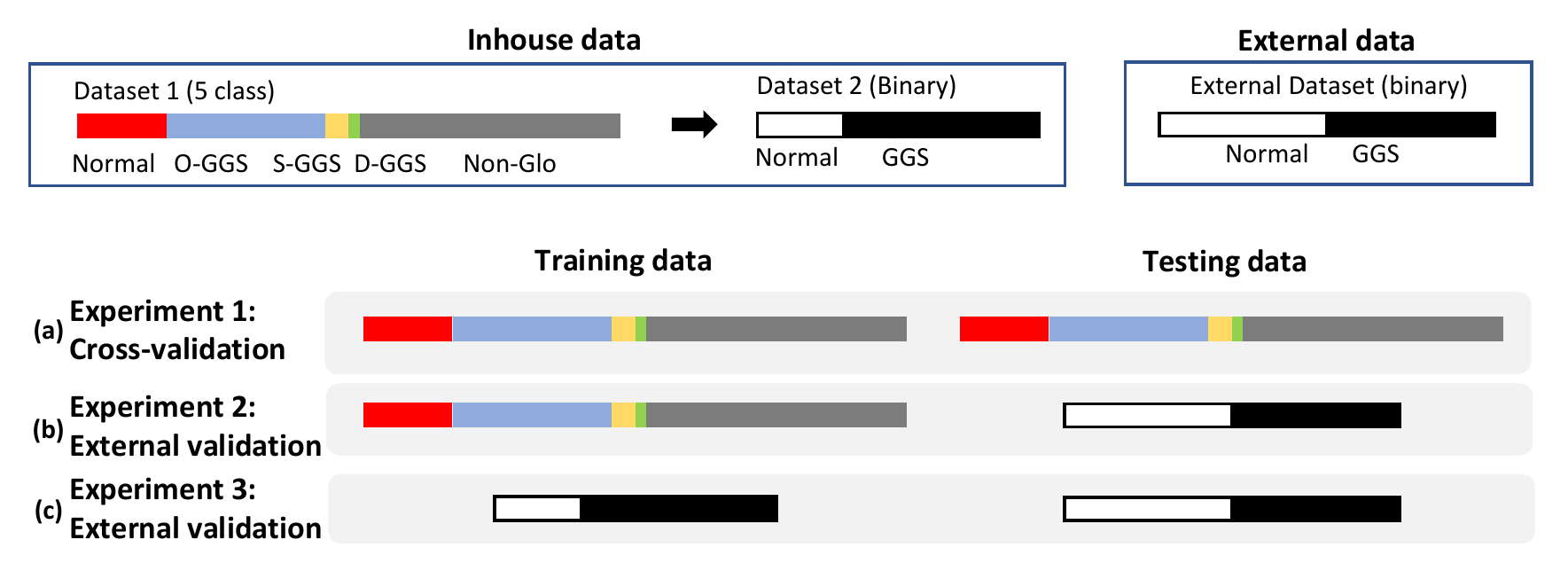}
\end{center}
   \caption{\textbf{Classification experiment setting}: (a) We evaluate different methods via the five-fold cross-validation using the in-house GGS dataset (multi-class). (b) An independent public dataset (binary) is employed to evaluate the trained models directly as an external validation. (c) The external validation is performed by retraining a binary classifier using different methods via the in-house GSS dataset (binary).}
   \label{fig:experiment}
\end{figure*}

Fig \ref{fig:experiment} shows a precise presentation for separate experiments.

All the models were trained and tested with five-fold cross-validation~\cite{refaeilzadeh2009cross}, where each fold was withheld as the testing data once~\cite{tang2013deep}. The remaining data for each fold was split as 75\% training data and 25\% validation data. Furthermore, to avoid data contamination, all glomeruli from the same patient were used either for training, validation, or testing. We employed a canonical design to prevent the data from leaking since the testing data were withheld from the training and validation data. 

In our study, we used the same hyperparameters (initial learning rate, batch size, and maximum number of epoch) and data augmentation strategies across all folds~\cite{wodzinski2021semi,perminov2021edge}. The batch size was determined as the maximum GPU memory limitation, while the initial learning rate (lr = 0.0001) was determined empirically by all validation data. Since the adaptive moment estimation (Adam) optimizer was used, even with the same initial learning rate, the underlying real learning rate was adjusted “fold-by-fold” as an adaptive optimization strategy. Via the five-fold cross-validation~\cite{tsamardinos2015performance}, the optimal epoch (model selection) of each fold was determined based on the validation performance as a “fold-by-fold” manner.

We adapted the SimSiam network~\cite{chen2020simple} as the baseline method of contrastive learning. Two random augmentations from the same image were used as training data. In self-supervised pretraining, 30,000 glomerulus pathologies from web mining were resized to $224\times224$ pixels for model training. For our web-mined glomerular dataset, 70\% of the images’ length-width ratio are within 4:3 or 3:4. Moreover, the random affine transformation was employed in the data augmentation, so that the images were resized to different aspect ratios, rotated with random degrees, and even with shearing before the downstream learning procedures. Therefore, we simply resized all images to $224\times224$ resolution images. This strategy has been widely used for large-scale image analysis in prior arts~\cite{krizhevsky2012imagenet,jayakumari2020automated,el2019deep,ali2019deep}. Two particular augmentation transforms-random resized crops and color distortions-are applied. In the classifier evaluation, only resizing to $224\times224$-pixel is applied in the augmentation. We used the momentum SGD as the optimizer. The Weight decay was set to 0.0001. The base learning rate was $lr=0.05$ and the batch size was 64. The learning rate was $lr\times$BatchSize$/256$, which followed a cosine decay schedule~\cite{loshchilov2017sgdr}.

We used ResNet-50 as the backbone in supervised training. The model was then trained using the Adam optimizer with a base learning rate of 1e-4. The optimizer learning rate followed (linear scaling~\cite{goyal2017accurate}) $lr\times$BatchSize$/256$. We used the focal loss \cite{lin2017focal} with $\gamma=2.5$ for unbalanced classes. Each fold was trained for 70 epochs, and the epoch with the best balanced accuracy score on the validation set was used for testing. 

To fully estimate the skill of the machine learning model on unseen data, all the models were trained and tested with five-fold cross-validation, where each fold was withheld as testing data once. The remaining data for each fold was split as 75\% training data and 25\% validation data. Furthermore, to avoid data contamination, all glomeruli from the same patient were used either for training, validation, or testing. 

To apply the self-supervised pre-training networks, we froze our pretrained ResNet-50 model by adding one extra linear layer which followed the global average pooling layer. When finetuning with the manually annotated glomeruli data, only the extra linear layer was trained. We used focal loss and the Adam optimizer to train the linear classifier with based (initial) learning rate of $lr$=30, weight decay=0, momentum=0.9, and batch size=64 (follows~\cite{chen2020exploring}). To evaluate the impact of training data size, we fine-tuned the classifier of our self-supervised learning model and the ImageNet pretrained model on 1, 5, 10, 25 and 100 percentages of annotated training data.

All experiments are performed on the same workstation, with a NVIDIA Quadro P5000 GPU of 16GB memory.

\subsubsection{Results}

\noindent\textbf{Cross Validation on In-house Dataset.} As shown in Table \ref{table:performance}, fine-tuning our pretrained SimSiam (Backbone:ResNet-50) on the same size of dataset achieved superior results as compared to training from scratch. Interestingly, our model also outperformed ResNet-50 models that were pretrained on ImageNet. Even Using 10\% of all available labeled data, our self-supervised ResNet50 model achieved the comparable performance with the fully supervised ResNet-50 models that used the entire labeled cohort. Moreover, the computational efficiency was also evaluated. Since contrastive learning only requires finetuning on the linear layers, it yielded less GPU memory (larger batch size) and higher computational efficiency (Table \ref{table:performance}). Moreover, our pretrained model also boosted accuracy upon the ImageNet pretrained ResNet-50.

\begin{figure*}
\begin{center}
\includegraphics[width=\linewidth]{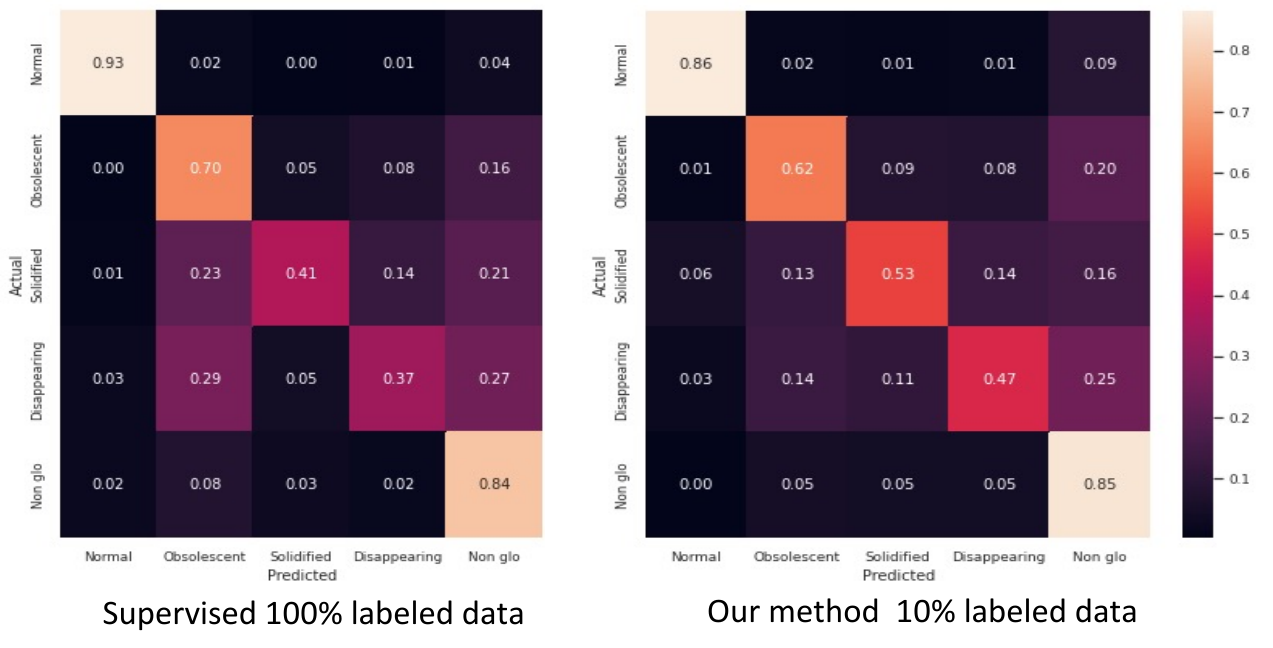}
\text{*Supervised method refers to training a ResNet-50 from scratch}
\end{center}
   \caption{Comparison between supervised method and our method }
\label{fig:cm}

\end{figure*}

The confusion matrices are shown in Fig.~\ref{fig:cm}. Our model achieved comparable performance as compared to the fully supervised models, while only using 10\% labeled data. Note that our approach had superior performance on the two categories (solidified and disappearing) which have smaller numbers of cases.

\begin{table*}
\centering{}
\caption{Performance of fine-grained GGS classification}
\begin{tabular}{cccccc}
\toprule
Model                & Unlabeled data & Labeled data & Balance acc & F1    & Time (s) \\
\midrule
\makecell{ResNet-50 \\ (train from scratch)}           & 0              & 100\%        & 65.0        & 61.7  &   159.8   \\
\midrule
\multirow{5}{*}{\makecell{ImageNet\\(finetune linear layers)}}  & \multirow{5}{*}{0}      & 100\%        & 62.0     & 51.9  &   43.7   \\
                              &        & 25\%    & 59.6     &50.5   &  36.6    \\
                              &       & 10\%     & 59.2      & 48.4  &   29.1 \\
                              &        & 5\%     &  58.0    &  46.1  &    28.7  \\
                              &         & 1\%    & 52.3     & 42.1   &   28.2   \\
\midrule
\multirow{5}{*}{\makecell{SimSiam\\(finetune linear layers)}}   & \multirow{5}{*}{30k}    & 100\%    & \textbf{68.0}    & \textbf{62.5} &   42.1  \\
                              &    & 25\%        &  66.8     &  60.8  &   35.2   \\
                              &     & 10\%       & 65.1        & 59.7  &  33.2    \\
                              &    & 5\%         & 64.4        &  54.8  &   29.8   \\
                              &   & 1\%          & 59.2        & 52.8   &  27.8   \\
\bottomrule
\end{tabular}
\label{table:performance}
\newline
\text{*ResNet-50: we trained the entire ResNet-50 from scratch with fully supervised learning.}
\newline
\text{*ImageNet: we only finetuned the linear layers of an ImageNet pretrained ResNet-50 model.}
\newline
\text{*SimSiam: we only finetuned the linear layers of a SimSiam pretrained ResNet-50 model.}
\newline
\text{*Time: indicates the average computational time per epoch.}
\newline
\end{table*}

\noindent\textbf{External Validation on Public Dataset.} For external validation, as shown in Table \ref{table:performance2}, trained with 5-classes labeled data, our models showed superior performance with a domain shift. The contrastive learning model achieved an AUC score of 94.5 when testing directly on the external dataset without any finetuning. Even the SimSiam model that previously finetuned on 10$\%$ of all available labeled data yielded better accuracy than the fully supervised ResNet-50 models.

In the binary setting (Table.~\ref{table:performance3}), the results yielded a similar trend when compared to the results from the 5-classes setting. Briefly, using only 10\% the labeled images, our model outperformed supervised training models that used all labeled images.

Note that finetuning the ImageNet pretrained model outperformed the training-from-scratch model in external validation, which is contrary to the results in cross validation. Such phenomenon might be caused by the domain shift in external validation. However, our method consistently outperformed those methods in different scenarios.

\begin{table*}[!htb]
\caption{Performance on external validation}
    \begin{minipage}{.5\linewidth}
    \subcaption{Train on 5-classes dataset}
    \medskip
    \resizebox{.95\textwidth}{!}{%
\begin{tabular}{ccc}
\toprule
Methods  & Labeled data   &AUC         \\
\midrule
\makecell{ResNet-50\\(train from scratch)}  & 100\%      &91.3          \\
\midrule
\makecell{ImageNet\\(finetune linear layers)} & 100\%     &92.4         \\ \midrule
\multirow{5}{*}{\makecell{SimSiam\\(finetune linear layers)}}    & 100\%         &\textbf{94.5}       \\ 
                            &  25\%         &  93.3               \\ 
                           &  10\%         & 92.6                 \\ 
                            &  5\%         & 92.2                 \\ 
                            &  1\%         &  89.9                \\ 
\bottomrule
\newline
\label{table:performance2}
\end{tabular}
    }

    \end{minipage} 
    \begin{minipage}{.5\linewidth}

    \subcaption{Train on binary dataset}
    \medskip

    \resizebox{.95\textwidth}{!}{%
\begin{tabular}{ccc}
\toprule
Methods  & Labeled data   &AUC         \\
\midrule
\makecell{ResNet-50\\(train from scratch)}  & 100\%      &93.3          \\
\midrule
\makecell{ImageNet\\(finetune linear layers)} & 100\%     &93.6         \\ \midrule
\multirow{5}{*}{\makecell{SimSiam\\(finetune linear layers)}}    & 100\%         &\textbf{95.2}       \\ 
                            &  25\%         &  94.8               \\ 
                           &  10\%         & 94.2                 \\ 
                            &  5\%         & 92.2                 \\ 
                            &  1\%         &  90.7                \\ 
\bottomrule
\label{table:performance3}
\end{tabular}    
     }

    \end{minipage}%
\newline
\text{*Definitions please see Table~\ref{table:performance}}
\end{table*}

\subsection{Glomerular Detection}

\begin{figure*}
\begin{center}
\includegraphics[width=\linewidth]{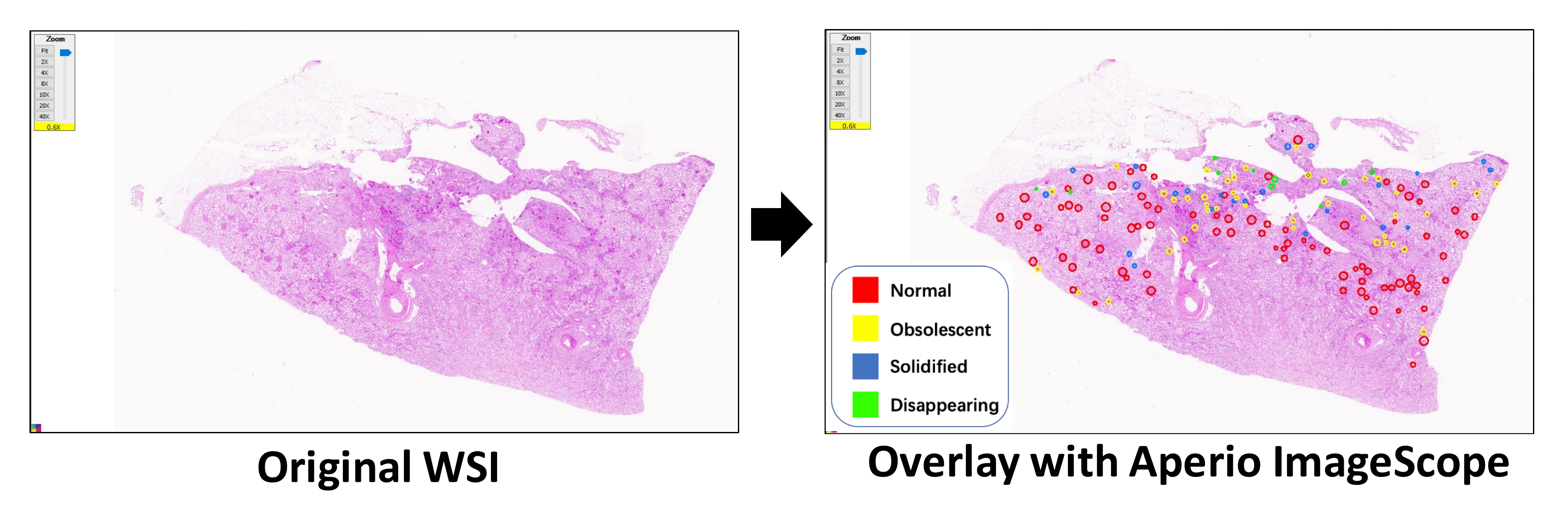}
\end{center}
   \caption{This figure presents the WSI-level multi-class circle glomerular detection results, which are visualized by ImageScope.}
\label{fig:result1}
\end{figure*}

\subsubsection{Experimental Design}

We employed the CircleNet pretrained model~\cite{yang2020circlenet,zhu2020easierpath} as the baseline model. In this study, we integrated the glomerular detection and classification as a unified framework to enhance the performance of object detection. Briefly, we introduced the ``non-glomerlar" class in the self-supervised $\mathsection$\textbf{Glomerular Classification} as a post-processing module to further filter out the false positive detection results. We report AP (average precision), AP50 and AP75 (average precision evaluated at 0.5 and 0.75 IoU threshold respectively), AP over small (APS), and medium (APM) objects, which shows the result of CircleNet detection~\cite{everingham2011pascal}.

\subsubsection{Results}
The results of glomerular detection was presented in Table.~\ref{detectresult} and Fig.~\ref{fig:result1}. By introducing the classifier as a false-positive filter, the detection performance in our Glo-In-One method achieved superior detection performance compared with original CircleNet method.

\subsection{Glomerular Segmentation}
\subsubsection{Experimental Design}
We conducted analyses on two of the widely used segmentation backbones (U-Net and DeepLab\_v3) on two unique resolutions (512$\times$512, 256$\times$256). Since the images from the previous stage are 256$\times$256 in the pipeline, we did experiments respectively on bilinear interpolation resizing and constant padding strategies. Mean and standard deviation of Dice similarity coefficients (DSC) were the primary statistics used to evaluate segmentation performance. The hyperparameters for training the U-Net and DeepLab\_v3 pipelines were 150 epochs, a batch size of 4 and a learning rate of 1e-4. An Adam Optimizer was used to adaptively alter the learning rate, with beta values ranging from 0.9 to 0.999.

\subsubsection{Results}
From the results (Table.~\ref{segresult}), our pipeline yielded decent DSC values across two backbone methods at different training scenarios, which is comparable to the prior arts~\cite{jha2021instance}. Among the benchmark methods, the DeepLab\_v3 with 512×512 image resolution is preferred.

\begin{figure}[!ht]
\begin{center}
\includegraphics[width=1\linewidth]{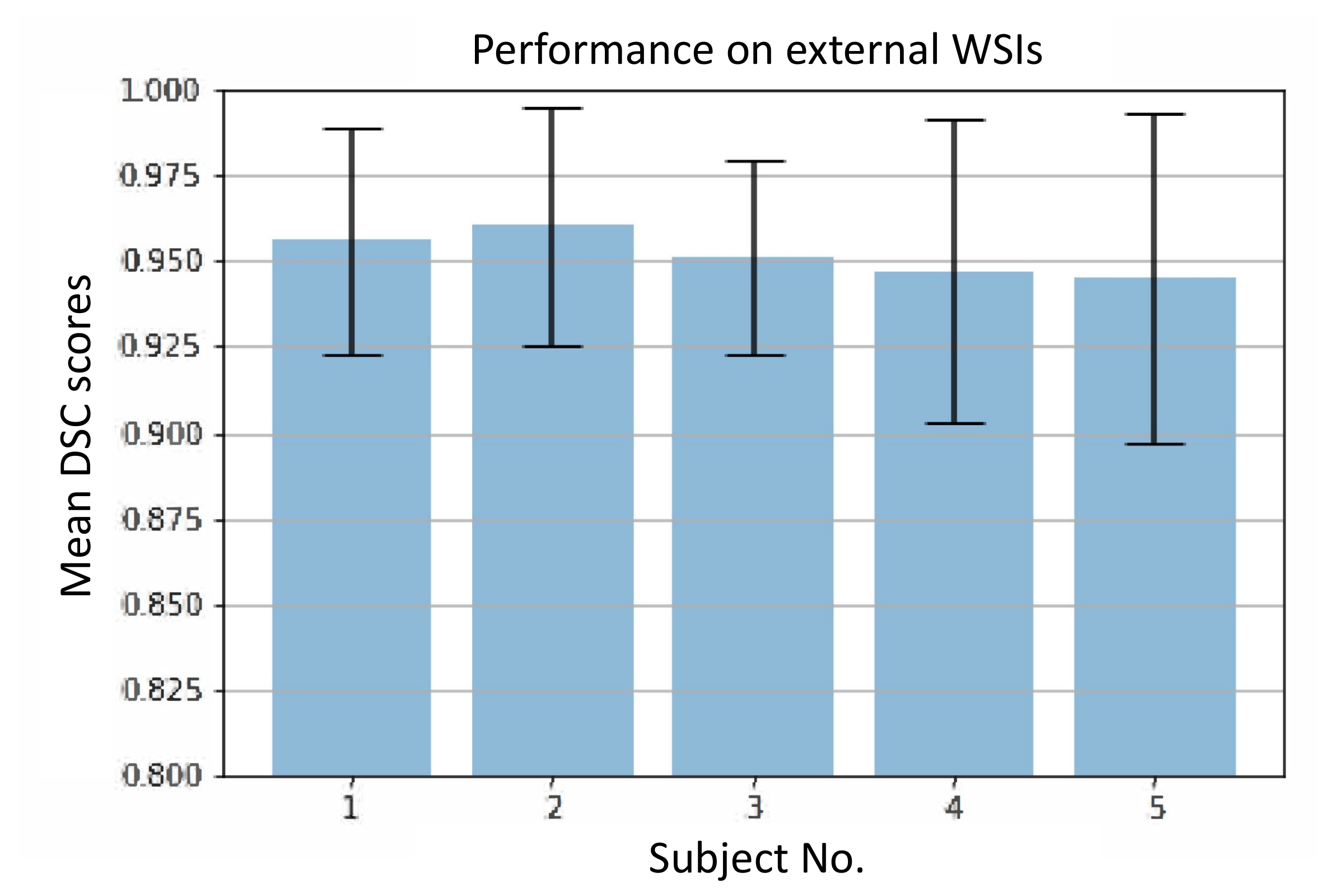}
\end{center}
   \caption{This figure presents the segmentation performance for glomerulus on external WSIs respectively. The bar plots show the mean values of Dice Similarity Coefficients (DSC) while the error bars represent the standard deviation.}
\label{fig:bar}
\end{figure}

Fig.~\ref{fig:bar} presents the segmentation performance at the individual WSIs level. The results show that the model achieves consistent performance across different slides.



\begin{table*}[!ht]
    \begin{minipage}{.45\linewidth}
    \subcaption{Detection Results}
    \medskip
    \resizebox{.95\textwidth}{!}{%
    \begin{tabular}{c c c c c c c }
         \toprule
          Method  & $AP$ & $AP_{50}$ & $AP_{75}$ & $AP_{S}$ & $AP_{M}$ \\
         \midrule
         CircleNet \quad  & 0.621 & 0.920 & 0.588 & 0.554 & 0.749 \\
         \makecell{Ours} \quad  & \textbf{0.627} & \textbf{0.951} & \textbf{0.604} & \textbf{0.573} & \textbf{0.758} \\
        \bottomrule
    
    \end{tabular}

    }
    \newline
    \text{*Ours: we used the CircleNet\_v2 model}
    \newline
    \text{with our second-stage classifier.}
    \newline
    \text{ }

    \label{detectresult}
    \end{minipage} 
    \begin{minipage}{.55\linewidth}

    \subcaption{Segmentation Results}
    \medskip
    \resizebox{.95\textwidth}{!}{%
    \begin{tabular}{c c c c }
    \hline
    Model & 512$\times 512^a$ & 512$\times512^b$ &256$\times$256 \\ 
    \hline
    U-Net & 
    0.934 ± 0.062 & 0.940 ± 0.029 & \textbf{0.947 ± 0.037} \\ 
    \hline
    DeepLab\_v3 & 
    \textbf{0.955 ± 0.045} & \textbf{0.950 ± 0.044} & 0.945 ± 0.047 \\
    \hline
    \end{tabular}
     }
    \newline
    \text{*512$\times 512^a$: constant padding from 256$\times$256}
    \newline
    \text{*512$\times 512^b$: bilinear resized from 256$\times$256}
    \newline
    \text{*256$\times 256$: original resolution}
    \label{segresult}
    \end{minipage}%
     \caption{\textbf{(a)} For Detection, we report AP (average precision), AP50 and AP75 (average precision evaluated at 0.5 and 0.75 IoU threshold respectively), AP over small (APS), and medium (APM) objects \textbf{(b)} For segmentation, we report mean and standard deviation of Dice similarity coefficients (DSC) to evaluate the two backbone methods.}
\end{table*}

\section{Discussion}
In this study, we introduced Glo-In-One, an integrated glomerular quantification open-source software toolkit for glomerular detection, segmentation, and lesion characterization. Only a single command line was required to perform all quantification tasks as a operating system agnostic Docker implementation. In this multitask system, we combined several well validated deep learning models and new self-supervised learning algorithms to enhance the reliability of the AI system. The proposed method would potentially be applied to the kidney transplant patients to achieve a more comprehensive pre-/post-operational evaluation with glomerulosclerosis characterization across all glomeruli. Moreover, the web image mining-based learning strategy might improve the diagnosis of rare pathological phenotypes by integrating the constantly updated knowledge from the community (e.g., from shared case studies on social media, new scientific publications, and new books).

Beyond releasing a holistic glomerular quantification tool, the core technical innovation was the self-supervised glomerular classification section, where we introduced the contrastive learning paradigm of fine grained GSS characterization via large-scale web image mining.The major focus of our method is to assess the feasibility and efficacy of incorporating the large-scale web mining data into computer-assisted nephropathological diagnosis to (1) improve the accuracy of the learning-based glomerulosclerosis characterization on unbalanced categories, and (2) enhance the generalizability of the AI models on unseen testing data.

This approach allows neural networks to learn more with fewer labels, smaller samples, or fewer trials. By taking advantages of these points, we come up with the idea of combining self-supervised learning and web image mining. Advanced detection algorithms and professional biomedical image database have disabused our worries about quality control when massively mining web images as training data. One promising future direction is to integrate self-supervised learning methods with web image mining in a poor annotation situation (in which only WSI-wise weak labels are provided) since our method does not require exhaustive tuning of hyperparameters and needs a smaller number of annotations.

In the classification experiment, the hyper-parameter tunning strategy in a cross-validation study is typically a “variance-bias” tradeoff. For example, as shown in the comprehensive study~\cite{tsamardinos2015performance}, cross-validation without hyper-parameter tunning yielded a lower variance but a higher bias compared with cross-validation with fold-by-fold hyper-parameter tunning. Therefore, we further conducted external validations (using an independent external cohort) to alleviate such concerns in cross-validation based evaluation, as a more rigor assessment.

From the results, the proposed strategy achieved superior performance in the task of GGS fine-grained classification, with fewer human efforts. By deploying our previously developed compound figure separation approach, we provided 30,000 web-mined unannotated glomerular images to pretrain the ResNet classification network via SimSiam contrastive learning. From the results, the GGS fine-grained classification model achieved superior performance when compared with baseline methods on both cross validation and external validation. However, the proposed method is still limited in its performance gain (mostly around 1\%). The recent advances in self-supervised learning have drawn significant attention towards the promising data efficiency and generalization ability. This approach assesses the feasibility of adapting such learning strategies to digital pathology with smaller scale annotated data and larger scale web mined data. One promising future direction is to further extend the proposed learning framework to the weakly supervised learning scenarios (in which only WSI-wise weak labels are provided). 

Several limitations and potential improvements for our work are as follows. First, integrating with a 3D registration algorithm would be an avenue for improvement for renal pathology as sequential cutting is typically involved~\cite{deng2021map3d}. With 3D quantification, we can potentially get more precise quantification of glomeruli, such as the average volume that represents the patients' degree of kidney sclerosis. Second, integrating interactive user quality assurance (QA) interface might enable more precise results with a ``human-in-the-loop" design. Third, from the results, the methods are limited by around 50\% of accuracy on GGS sub-classes classification, another potential improvement in terms of the performance could be to build up a more class-balanced fine grained GGS dataset. The binary classification cohort was used as our external validation data. Another external validation cohort with multiple GSS types would lead to a more comprehensive external validation. However, to our best, such a dataset is not publicly available to further exam the generalizability of our multi-class model.

\section{Conclusions}
In this paper, we develop and release a holistic Glo-In-One open-source toolkit to provide holistic glomerular detection, segmentation, and lesion characterization. The containerized machine learning system process gigapixel renal WSIs in a fully automated manner with a single command line, which is user-friendly for non-technical users. Our technical contribution focuses on combining cost-efficient and large-scale web image mining with self-supervised algorithms to achieve superior performance in glomerular detection and fine grained GSS classification. The results demonstrate that our self-supervised ResNet50 achieves superior performance as compared with a standard supervised ResNet-50 using all labeled images.

\section*{Acknowledgment} 
This work was supported by NIH NIDDK DK56942(ABF). 


\ifCLASSOPTIONcaptionsoff
  \newpage
\fi



\bibliographystyle{IEEEtran}
\bibliography{main}
%







\end{document}